\documentclass[letterpaper]{article} 
\usepackage{aaai2027}  
\usepackage[hyphens]{url}  
\usepackage{graphicx} 
\usepackage{natbib}  
\usepackage{caption} 
\usepackage{algorithm}
\usepackage{algorithmic}

\usepackage{newfloat}
\usepackage{listings}
\DeclareCaptionStyle{ruled}{labelfont=normalfont,labelsep=colon,strut=off} 
\floatstyle{ruled}
\newfloat{listing}{tb}{lst}{}
\floatname{listing}{Listing}

\usepackage{booktabs}

\usepackage{amsmath}
\usepackage{amssymb}
\usepackage{array}
\usepackage{multirow}
\usepackage{placeins}

\title{VLALight: Lightweight Vision-Language-Action Models for Emergency-Aware Traffic Signal Control}
\author{
   Kemou Jiang\textsuperscript{\rm 1},
   Maonan Wang\textsuperscript{\rm 2,3},
   Xingchen Zou\textsuperscript{\rm 4},
   Jiayue Zhu\textsuperscript{\rm 1},
   Yuhang Fu\textsuperscript{\rm 1},
   Sicheng Wang\textsuperscript{\rm 1},
   Xi Chen\textsuperscript{\rm 3},
   Yirong Chen\textsuperscript{\rm 5},
   Zhiyong Cui*\textsuperscript{\rm 1}
}
\affiliations{
    \textsuperscript{\rm 1}School of Transportation Science and Engineering, Beihang University, Beijing, China\\
    \textsuperscript{\rm 2}The Chinese University of Hong Kong (Shenzhen), Shenzhen, China\\
    \textsuperscript{\rm 3}The Chinese University of Hong Kong, Hong Kong SAR, China\\
    \textsuperscript{\rm 4}The Hong Kong University of Science and Technology (Guangzhou), Guangzhou, China\\
    \textsuperscript{\rm 5}Shanghai Artificial Intelligence Laboratory, Shanghai, China\\
    *Corresponding author
}

\begin{document}

\maketitle

\begin{abstract}
Traffic signal control (TSC) is essential for mitigating urban congestion. Recent advances in vision-language models (VLMs) enable richer interpretation of intersection scenes, opening new opportunities for visual-context-aware TSC. However, the loose coupling and repeated information conversion between modules can lead to the loss of fine-grained visual details, while sequential inference introduces substantial latency. To address these limitations, we propose VLALight, a lightweight end-to-end vision-language-action framework that directly maps intersection observations and signal-phase information to discrete signal actions. To handle the multi-view nature of TSC, VLALight combines multiple directional camera views into a unified visual input and uses textual instructions to establish their correspondence with traffic movements and signal phases. This design enables direct action prediction with a compact 0.5 B-parameter model, without intermediate image-to-text descriptions or handcrafted traffic-state representations. Experiments show that VLALight delivers the best emergency-vehicle service of all compared methods, reducing pooled emergency waiting time by 21.1\% over the cascaded VLMLight while running in real time on local hardware and generalizing to unseen intersection topologies and traffic-flow patterns.
\end{abstract}

\section{Introduction}

Urban traffic congestion remains a major challenge in modern cities. Intersections are the bottlenecks of road networks, and their signal strategies largely determine network-wide efficiency and safety. Traffic signal control (TSC) methods have evolved from fixed-time plans to adaptive rule-based strategies such as SOTL~\cite{gershenson2005self} and Max Pressure~\cite{varaiya2013maxpressure} and, more recently, deep reinforcement learning (DRL)~\cite{wei2021recent}. These controllers react to real-time measurements and perform well under steady demand, but they almost universally represent the traffic state with compact hand-crafted statistics such as queue lengths, lane occupancy, and vehicle counts. Because such statistics discard fine-grained spatial and interaction information, these controllers can misjudge the situation or degrade sharply in abnormal scenarios such as accidents, sudden congestion, or emergency-vehicle priority requests, where scene understanding matters most. Large language models (LLMs) and vision-language models (VLMs) have recently been introduced to inject semantic understanding into TSC~\cite{lai2024llmlight,wang2025vlmlight,zou2025trafficr1}, yet they adopt a cascaded ``perception, reasoning, and control'' pipeline in which traffic images are first converted into textual descriptions, over which an LLM then reasons to produce control suggestions; the vision-to-text conversion again compresses spatial detail, and multi-stage inference adds latency that is problematic for real-time control. Both lines of work thus share a common bottleneck: \emph{the visual scene is compressed before decisions are made}.

Vision-language-action (VLA) models offer a direct way to remove this bottleneck. By tokenizing actions and treating control as a generative prediction problem, models such as RT-2~\cite{zitkovich2023rt2}, Octo~\cite{ghosh2024octo}, and OpenVLA~\cite{kim2024openvla} unify perception, language understanding, and action execution in a single end-to-end network, preserving full visual detail up to the decision layer while inheriting web-scale perceptual and linguistic priors. These properties are exactly what emergency-aware traffic control demands: the controller must see the whole scene, understand topology and intent expressed in language, and act directly. Yet VLA research has so far focused on robot manipulation~\cite{yu2025efficient,ye2025vlar1,zhang2026vlm4vla}; in this work, we bring the VLA paradigm to traffic signal control.

In TSC scenarios, however, VLA models face several new challenges. \emph{First, the observation structure differs.} A robot policy consumes a temporal stream from a single ego-centric view, whereas a traffic controller must process multiple directional views at once, and each phase decision affects only the traffic movements visible in some of the views; the model must therefore relate the views to each other and parse the lane-level structure within each view. \emph{Second, intersection topologies vary.} Real intersections differ in shape, topology, and lane count, yet a single policy must serve all of them, and ideally unseen ones as well, which places a higher demand on generalization. \emph{Third, the control objective is event-driven.} Beyond serving routine traffic efficiently, the controller must also spot special vehicles such as ambulances and fire engines in dense ordinary traffic, rare events that demand an immediate shift of phase priority in their favor; such long-tail events are precisely where compact state statistics and efficiency-driven training are weakest. Real-time operation imposes a further practical constraint on model size and adaptation cost.

We address these challenges with VLALight, a lightweight end-to-end VLA framework for intersection signal control. Because real surveillance footage paired with control labels is difficult to obtain at scale, we first build a simulation-grounded data-generation pipeline that replicates the lane-level topology of real intersections, designs diverse traffic-flow patterns with injected emergency vehicles, and renders high-fidelity camera images with verified expert labels. On the modeling side, VLALight stitches the directional camera views of an intersection into a single panoramic observation, mirroring how a human operator reads a surveillance wall, and pairs it with a synthesized textual instruction that establishes the correspondence between camera views, traffic movements, and signal phases. Conditioned on these inputs, a compact 0.5B multimodal backbone, adapted with LoRA through the VLA-Adapter paradigm~\cite{wang2025vlaadapter} and augmented with a small set of learnable phase queries, directly decides the next phase in a single forward pass, with no intermediate image-to-text descriptions or handcrafted traffic-state representations at any stage. Figure~\ref{fig:framework} illustrates the overall architecture. Our main contributions are:

\begin{figure}[t!]
\centering
\includegraphics[width=\columnwidth]{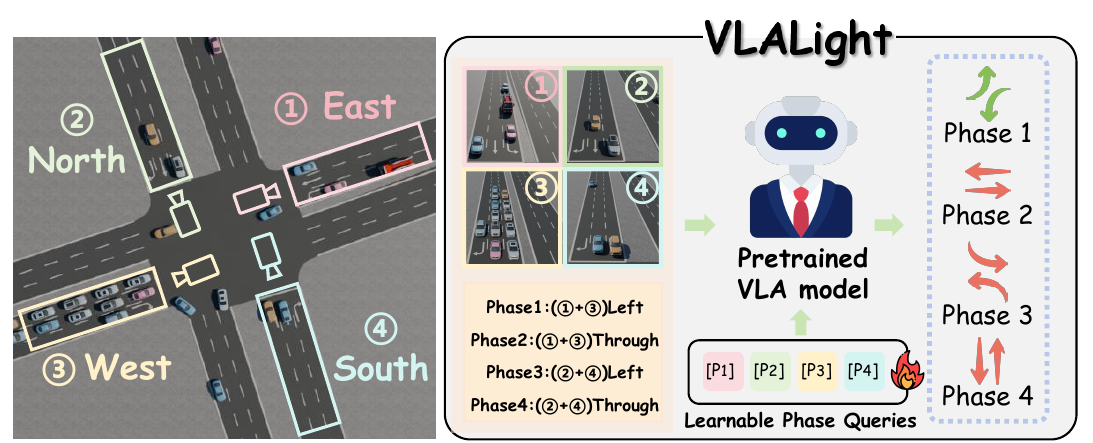}
\caption{Overview of VLALight.}
\label{fig:framework}
\end{figure}

\begin{itemize}
    \item We build a simulation-grounded visual data-generation pipeline for TSC that replicates real-world intersection topologies, synthesizes diverse traffic flows with injected emergency vehicles, collects expert labels under human verification, and renders high-fidelity multi-view images, yielding a verified multi-intersection dataset for emergency-aware signal control.
    \item We design a unified multi-directional visual representation and a TSC-native phase-classification action space that enable direct action prediction with a compact multimodal model.
    \item We propose VLALight, a lightweight end-to-end vision-language-action framework for traffic signal control that maps intersection observations and signal-phase information directly to discrete signal actions, running in real time on local hardware.
\end{itemize}

\section{Related Work}

\textbf{Traffic signal control.} Control strategies have progressed from fixed-time plans to adaptive rule-based strategies and, more recently, learning-based policies. Fixed-time controllers derive cycle lengths and green splits from historical traffic statistics~\cite{webster1958traffic}, whereas adaptive methods such as SOTL~\cite{gershenson2005self} and Max Pressure~\cite{varaiya2013maxpressure} react to real-time detector measurements. Deep reinforcement learning has further advanced the field, spanning value-based methods~\cite{clifton2020qlearning,hasselt2016deep,huang2018vdd3qn}, policy-gradient algorithms~\cite{schulman2017proximal}, and multi-agent coordination frameworks~\cite{wei2019colight,chen2020mplight,wei2018intellilight,wei2019presslight,oroojlooy2020attendlight,wang2024unitsa}.

\begin{figure*}[t]
\centering
\includegraphics[width=\textwidth]{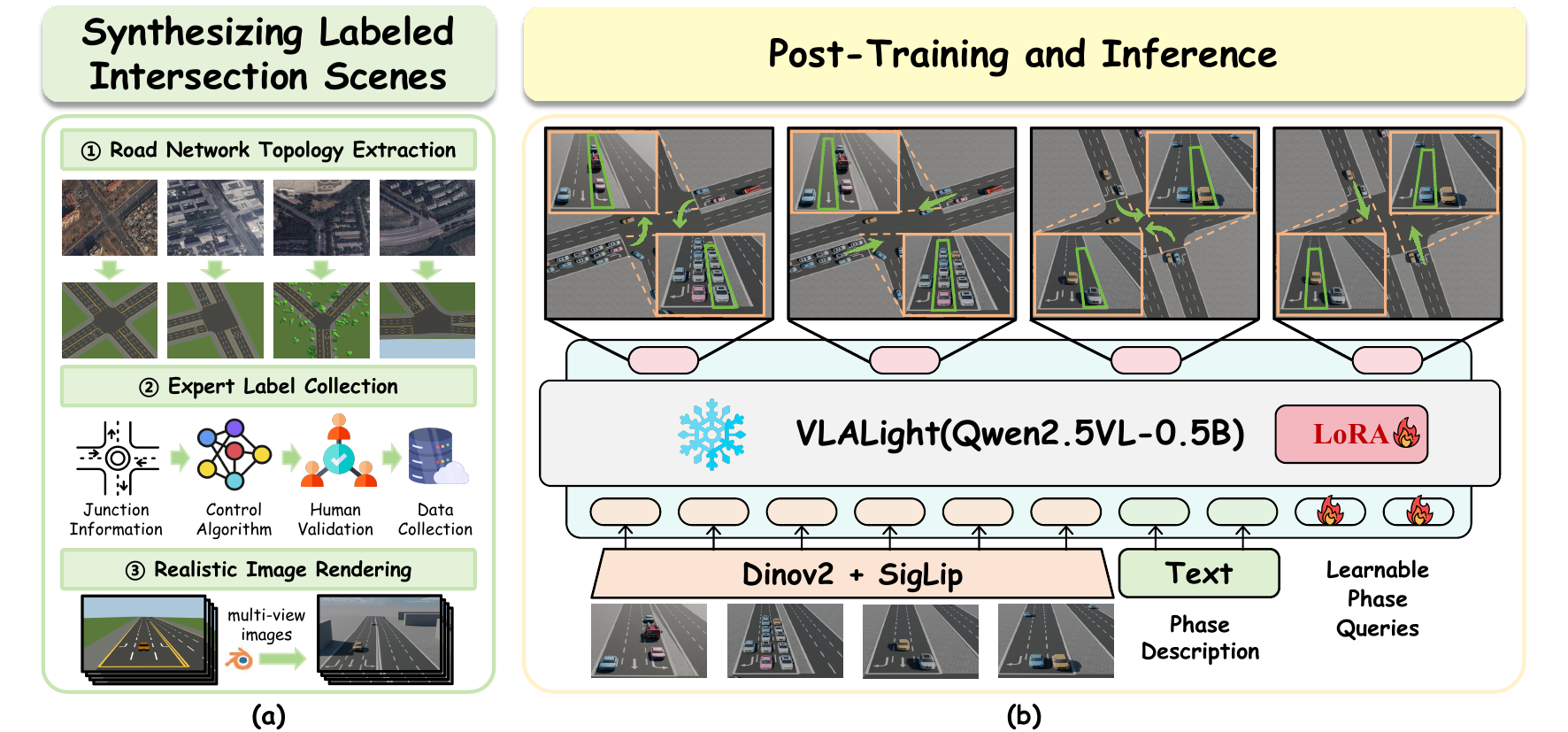}
\caption{Architecture of VLALight. (a) The simulation-grounded traffic data-generation pipeline: road network topology extraction, traffic flow design and expert label collection, and realistic intersection image rendering. (b) Post-training and inference: the directional camera images and the text instruction are encoded by the frozen DINOv2 and SigLIP visual encoders and the compact VLM backbone, which directly outputs the next signal phase.}
\label{fig:method}
\end{figure*}

\textbf{LLMs and VLMs for traffic control.} To bring semantic reasoning into TSC, recent studies have turned to LLMs and VLMs~\cite{nie2025exploring}. Early systems encode structured detector states as text and prompt an LLM for decisions~\cite{lai2024llmlight,masri2024large,wang2024llmassisted,da2024prompt}, while more recent designs add visual perception, coupling visual encoders with language reasoning for signal control~\cite{wang2025vlmlight,zou2025trafficr1}.

\textbf{Vision-language-action models.} VLA models offer an end-to-end alternative by casting control as generative action prediction. RT-1~\cite{brohan2023rt1}, RT-2~\cite{zitkovich2023rt2}, Octo~\cite{ghosh2024octo}, and OpenVLA~\cite{kim2024openvla} demonstrated this paradigm at scale in robotic manipulation, and subsequent work has improved efficiency~\cite{pertsch2025fast,yu2025efficient}, reasoning~\cite{zhao2025cotvla,ye2025vlar1,zhang2026vlm4vla}, action representations~\cite{black2024pi0,chi2023diffusionpolicy,zhao2023act}, and safety~\cite{zhang2025safevla}. A parallel line of compact architectures, including RoboFlamingo~\cite{li2024roboflamingo}, UniVLA~\cite{bu2025univla}, and OpenVLA-OFT~\cite{kim2025openvlaoft}, reduces model size or latency, and a recent survey~\cite{yu2025efficient} provides a comprehensive overview of this fast-growing literature.

\section{Method}

This section first formulates emergency-aware traffic signal control as a vision-language-action decision problem, then describes the simulation-grounded pipeline that generates the training data, and finally presents the VLALight policy.

\subsection{Problem Formulation}

An intersection connects incoming and outgoing approaches, each carrying one or more lanes; a \emph{movement} carries vehicles from an incoming approach to an outgoing one, and movements are grouped into \emph{phases}, each a set of non-conflicting movements that receive the green signal simultaneously. We consider intersection control with a phase set of size $K$, and time is discretized into decision steps of length $\Delta t$. At each step $t$, the agent observes a stitched surveillance image $I_t\in\mathbb{R}^{H\times W\times3}$ covering all approaching directions, together with a natural-language instruction $L$ that describes the intersection topology and the meaning of each signal phase, and directly outputs a discrete phase $a_t\in\{0,\dots,K{-}1\}$, equivalently a one-hot vector $\mathbf{a}_t\in\{0,1\}^K$. The goal is to learn a policy $\pi_\theta(a_t\mid I_t,L)$ that minimizes the average travel and waiting times of all vehicles while remaining responsive to emergency-vehicle priority requests. The following sections present the two designs that adapt the VLA paradigm to this task, namely a multi-directional visual representation and a TSC-native action space, together with the overall VLALight architecture illustrated in Figure~\ref{fig:method}; a detailed comparison with VLA models in other domains is deferred to the supplementary material (Section~E).

\subsection{Synthesizing Labeled Intersection Scenes from Real Topology}

Real intersection footage paired with control labels is difficult to obtain at scale: raw surveillance video raises privacy concerns and is rarely released, and the signal decisions that were executed while the cameras rolled are equally hard to access, so usable supervision labels do not exist in the real world. We therefore generate training data in simulation while anchoring every stage to the real world, through three steps: replicating real road-network topology, designing traffic flows and collecting expert labels, and rendering realistic camera images (details in the supplementary material, Section~A).

\subsubsection{Road Network Topology Extraction}

To make the simulated scenes faithful to real deployment conditions, we replicate six real-world intersections in the TransimHub simulation platform~\cite{wang2025transimhub} with their lane-level topology exactly preserved. For each site, we first obtain the raw road-network data from OpenStreetMap, and then refine it against Google satellite imagery and street-view photos, correcting lane counts, lane functionalities, and traffic organization until the simulated network matches reality. The selected intersections cover both four-way cross and T-junction layouts and differ in lane count and phase design, so that the collected data capture diverse signal-control scenarios instead of overfitting to one specific intersection layout.

\subsubsection{Expert Label Collection}

For each intersection we design a \emph{regular flow} that reproduces a typical day with moderate, smoothly varying volumes, and collect the expert demonstrations under it; additional demand patterns used only for evaluation are introduced in the Experiments section. We inject emergency vehicles (police cars, ambulances, and fire engines) with predefined types, entry times, and paths. A control label is then attached to every decision step according to the scenario. Under routine demand, where the objective is to discharge queues and minimize congestion, each step is labeled by a maximum-pressure expert~\cite{varaiya2013maxpressure}, which selects the phase with the highest pressure, defined as the aggregated difference between incoming and outgoing vehicle counts over the movements it serves. Max pressure is blind to emergency vehicles, so whenever an emergency vehicle is present, we instead query a multimodal LLM with the visual scene to select the phase that serves its movement. All labels are finally verified by traffic signal control experts, yielding the dataset $\mathcal{D}=\{(I,L,\mathbf{a}^*)\}$ of stitched images, instructions, and one-hot expert actions. The demonstrations thus teach the policy both to balance pressure and to prioritize emergency vehicles.

\subsubsection{Realistic Image Rendering}

Raw simulator frames suffer from low resolution, blurry textures, and missing detail, and policies trained on them transfer poorly to real scenes. We therefore render every scene with Blender's high-fidelity rendering engine, producing realistic lighting, materials, and vehicle appearance that narrow the sim-to-real visual gap. Each intersection is instrumented with up to four cameras, one per approach, placed at roadside surveillance viewpoints rather than a bird's-eye view, matching how intersections are actually monitored.

\subsection{Post-Training and Inference}

VLALight follows the VLA-Adapter paradigm~\cite{wang2025vlaadapter}: a frozen pretrained VLM is augmented with a small set of lightweight trainable modules, among them a set of $N_a$ learnable phase queries $\mathbf{Q}=\{\mathbf{q}_j\}_{j=1}^{N_a}$, and predicts the next phase in a single forward pass. To bridge the domain differences identified above, we first present the two key designs, the multi-directional visual representation and the TSC-native action space, and then walk through the complete policy-learning pipeline.

\subsubsection{Mutli-view Visual}

A traffic controller sees the world very differently from a robot. In practice, intersections are rarely observable from a bird's-eye view; they are monitored by roadside cameras, one per approach, and a signal decision must coordinate traffic movements that are visible across different views. More importantly, choosing a phase is a trade-off among competing directions: granting green time to the movements in one view withholds it from all others, so a sound decision must weigh the demand visible in every view against each other rather than reading any view in isolation. This is why human operators watch the camera feeds arranged side by side on a surveillance wall: only a joint view of all directions supports such cross-view weighing. We give the model the same joint view, composing a single panoramic observation by stitching the directional views into a $2\times2$ grid:
\begin{equation}
    I_t = \operatorname{Stack}_{2\times2}\big(\{C_t^{(i)}\}_{i=0}^{3}\big) \in \mathbb{R}^{H\times W\times3},
\end{equation}
where $C_t^{(i)}\in\mathbb{R}^{H_c\times W_c\times3}$ is the $i$-th camera view and missing views are zero-padded. Encoding the views jointly is essential here: the cross-view trade-off above must be made within one shared representation, and a compact backbone has limited ability to relate separately encoded views. Following multi-camera VLAs that place all camera streams in one sequence before the language model, the stitched image is resized and normalized as a whole and encoded in a single pass by two frozen pretrained encoders, with DINOv2 contributing self-supervised geometric features and SigLIP providing language-aligned semantic features~\cite{oquab2023dinov2,zhai2023siglip,radford2021clip}. With the ViT/14 backbones at their native input resolution, the stitched image yields $256$ patch tokens arranged in a $16\times16$ spatial grid, and the two encoders' features are concatenated patch-wise. Because the grid inherits the $2\times2$ layout of the stitched image, each $8\times8$ token block keeps its directional identity, so the instruction can ground every token block to a direction, its movements, and its phase. All $256$ tokens enter the language model as one sequence within a single forward pass of the 0.5B backbone, where self-attention contextualizes every visual token by all views at once:
\begin{equation}
\label{eq:vltokens}
\begin{gathered}
    \mathbf{v} = \big[\,\mathbf{v}^{\text{dino}} \,\|\, \mathbf{v}^{\text{siglip}}\,\big], \quad
    \mathbf{Z} = \operatorname{MLP}_{\text{proj}}(\mathbf{v}),
\end{gathered}
\end{equation}
where $\mathbf{v}^{\text{dino}}$ and $\mathbf{v}^{\text{siglip}}$ are the patch features of the stitched image from the two encoders, $\|$ denotes concatenation along the feature dimension, and $\mathbf{Z}$ holds the $N_p$ visual tokens projected to the LLM hidden size. The instruction $L$ is a concise textual description of the intersection, covering the approaching direction monitored by each camera and the traffic movement enabled by each phase index, inserted into a fixed prompt template; exposing topology through text rather than architecture lets a single policy serve intersections with different layouts and lane counts by only changing the instruction.

\subsubsection{TSC-Native Action Space}

In signal control, the decision itself is the choice of a phase: at each step the controller selects the next phase from a small discrete set rather than emitting continuous signals. We therefore cast the decision as phase classification, formulated as token prediction in the same representation space as vision and language~\cite{zitkovich2023rt2,kim2024openvla}. Concretely, each action dimension is tied to a reserved token from the end of the LLM vocabulary~\cite{yang2024qwen25}, and during training these positions are occupied by a shared set of learnable phase queries, decoupling the action representation from the tokenizer vocabulary. The action head maps the phase-query hidden states to the $K$ phase logits $\hat{\mathbf{a}}\in\mathbb{R}^K$, and the predicted phase is $\hat{p}_t=\operatorname*{argmax}_{j}\,\hat{a}_{t,j}$. At deployment, the prediction passes through a minimum-green dwell constraint: once activated, a phase is held for a minimum duration before any switch is permitted, mirroring real signal-timing rules and preventing the policy from oscillating between locally attractive phases across decision steps.

\subsubsection{Policy Learning}

Algorithm~\ref{alg:algorithm} gives the complete training and inference procedure, and all hyperparameters are listed in the supplementary material (Section~B.2). A frozen pretrained VLM, comprising dual vision encoders and a compact language model, is augmented with four lightweight trainable modules: a multilayer projector, LoRA adapters, the phase queries $\mathbf{Q}$, and a residual-MLP action head. At each decision step, the projected visual tokens $\mathbf{Z}$ are placed right after the beginning-of-sequence embedding so that every later position can attend to the full visual context under the causal mask, followed by the instruction tokens and the phase queries:
\begin{multline}
    \mathbf{E} = [\mathbf{e}_{\text{BOS}}; \mathbf{Z}_1; \dots; \mathbf{Z}_{N_p}; \mathbf{e}_1; \dots; \mathbf{e}_{N_L}; \\
    \mathbf{q}_1; \dots; \mathbf{q}_{N_a}; \mathbf{e}_{\text{EOS}}],
\end{multline}
where $\mathbf{e}_{\text{BOS}}$ and $\mathbf{e}_{\text{EOS}}$ are the beginning- and end-of-sequence embeddings, $\mathbf{Z}_1,\dots,\mathbf{Z}_{N_p}$ are the $N_p$ projected visual tokens, $\mathbf{e}_1,\dots,\mathbf{e}_{N_L}$ are the embeddings of the $N_L$ instruction tokens, and $\mathbf{q}_1,\dots,\mathbf{q}_{N_a}$ are the $N_a$ learnable phase queries; only the phase-query positions are supervised. Let $\mathbf{H}^{(l)}$ denote the hidden states of the $l$-th LLM layer. A zero-initialized action state $\mathbf{x}^{(0)}$ is refined through residual blocks paired one-to-one with the LLM layers; each block performs a tri-branch attention that fuses, under a single softmax, keys and values from three sources (the action state itself, the phase-query hidden states, and the visual hidden states) with rotary position embeddings applied within each branch:
\begin{equation}
\label{eq:tribranch}
\begin{aligned}
    \operatorname{Attn}(\mathbf{x}) = \operatorname{softmax}\!\left(\frac{\mathbf{q}\,[\mathbf{k}_x;\,\mathbf{k}_a;\,\tanh(g)\,\mathbf{k}_v]^{\top}}{\sqrt{d_h}}\right)\\
    \cdot\,[\mathbf{v}_x;\,\mathbf{v}_a;\,\mathbf{v}_v],
\end{aligned}
\end{equation}
where $\mathbf{q}$ is a linear projection of the action state $\mathbf{x}$; $(\mathbf{k}_x,\mathbf{v}_x)$, $(\mathbf{k}_a,\mathbf{v}_a)$, and $(\mathbf{k}_v,\mathbf{v}_v)$ are the key-value pairs computed from the action state, the phase-query hidden states, and the visual hidden states, respectively; and $d_h$ is the head dimension. The visual branch is gated by $\tanh(g)$ with a learnable scalar $g$ initialized to zero, so the head relies on action features early in training and gradually incorporates visual context as the gate opens. After a residual connection and a feed-forward layer, a final linear projection maps the refined action state to the phase logits,
\begin{equation}
    \hat{\mathbf{a}} = \mathbf{W}_{o}\,\mathbf{x}^{(L)} + \mathbf{b}_{o},
\end{equation}
where $\mathbf{x}^{(L)}$ is the action state after the $L$ residual blocks, and $\mathbf{W}_{o}$ and $\mathbf{b}_{o}$ are the output-projection weight and bias. Training is behavioral cloning of the verified expert labels: given the one-hot ground-truth action $\mathbf{a}^*$, we minimize the cross-entropy over the $K$ phases,
\begin{equation}
    \mathcal{L}_{\text{CE}} = -\sum_{j=0}^{K-1} a_j^* \log \frac{\exp(\hat{a}_j)}{\sum_{k=0}^{K-1}\exp(\hat{a}_k)},
\end{equation}
updating only the lightweight modules $\theta=\{\theta_{\text{proj}},\theta_{\text{LoRA}},\mathbf{Q},\theta_{\text{head}}\}$ while the vision encoders and the base LLM remain frozen, preserving the visual-linguistic priors of large-scale pretraining~\cite{liu2023llava,li2023blip2,radford2021clip} with a minimal number of trainable parameters.

\begin{algorithm}[t]
\caption{VLALight Training and Inference}
\label{alg:algorithm}
\textbf{Input}: Expert demonstrations $\mathcal{D}=\{(I,L,\mathbf{a}^*)\}$ (stitched image, instruction, one-hot expert action)\\
\textbf{Parameters}: Vision encoder $E$, LLM $f_{\text{LLM}}$, projector $g_{\text{proj}}$, action head $f_{\text{head}}$, phase queries $\mathbf{Q}$, adapter parameters $\theta$\\
\textbf{Output}: Policy $\pi_\theta$
\begin{algorithmic}[1]
\STATE Initialize $\theta \gets \{\theta_{\text{proj}},\theta_{\text{head}},\mathbf{Q},\theta_{\text{LoRA}}\}$
\STATE Freeze $E$ and $f_{\text{LLM}}$
\WHILE{not converged}
    \STATE Sample a batch $\{(I^i,L^i,\mathbf{a}^{*i})\}_{i=1}^{B}$ of size $B$ from $\mathcal{D}$
    \STATE $\mathbf{v}^i = E(I^i)$
    \STATE $\mathbf{Z}^i = g_{\text{proj}}(\mathbf{v}^i)$
    \STATE $\mathbf{E}^i = [\mathbf{e}_{\text{BOS}}; \mathbf{Z}^i; \operatorname{Tokenizer}(L^i); \mathbf{Q}; \mathbf{e}_{\text{EOS}}]$
    \STATE $\hat{\mathbf{a}}^i = f_{\text{head}}\big(f_{\text{LLM}}(\mathbf{E}^i;\theta)\big)$
    \STATE Update $\theta$ by minimizing $\mathcal{L}_{\text{CE}}(\hat{\mathbf{a}}^i, \mathbf{a}^{*i})$
\ENDWHILE
\FOR{each evaluation step $t$}
    \STATE Observe $(I_t, L_t)$
    \STATE $\mathbf{E}_t = [\mathbf{e}_{\text{BOS}}; E(I_t); \operatorname{Tokenizer}(L_t); \mathbf{Q}; \mathbf{e}_{\text{EOS}}]$
    \STATE $\hat{\mathbf{a}}_t = f_{\text{head}}\big(f_{\text{LLM}}(\mathbf{E}_t; \theta)\big)$
    \STATE Execute $\hat{p}_t = \operatorname*{argmax}_{j}\, \hat{a}_{t,j}$ subject to the minimum-green dwell constraint
\ENDFOR
\STATE \textbf{return} $\pi_\theta$
\end{algorithmic}
\end{algorithm}

\section{Experiments}

\begin{table*}[t]
\centering
\small
\renewcommand{\arraystretch}{1.3}
\setlength{\tabcolsep}{0.8pt}
\begin{tabular}{@{}c|c|cc|cc|cc|cc|cc|cc@{}}
\hline
\multirow{2}{*}{Type} & \multirow{2}{*}{Method} & \multicolumn{2}{c|}{Beihuan} & \multicolumn{2}{c|}{Beishahe} & \multicolumn{2}{c|}{Changjianglu} & \multicolumn{2}{c|}{Gaojiaoyuan} & \multicolumn{2}{c|}{Pinganli} & \multicolumn{2}{c}{Zhijingdao} \\
\cline{3-14}
 &  & ETT$\downarrow$ & EWT$\downarrow$ & ETT$\downarrow$ & EWT$\downarrow$ & ETT$\downarrow$ & EWT$\downarrow$ & ETT$\downarrow$ & EWT$\downarrow$ & ETT$\downarrow$ & EWT$\downarrow$ & ETT$\downarrow$ & EWT$\downarrow$ \\
\hline
\multirow{4}{*}{Trad.} & FixTime & \textbf{59$\pm$19}\smash{\textsuperscript{$\ddagger$}} & \textbf{19$\pm$12}\smash{\textsuperscript{$\dagger$}} & 44$\pm$2.9 & 12$\pm$2.6 & \textbf{37$\pm$4.1}\smash{\textsuperscript{$\dagger$}} & \textbf{1.6$\pm$2.8}\smash{\textsuperscript{$\dagger$}} & 57$\pm$6.7 & 22$\pm$7.2 & 51$\pm$7.6 & 16$\pm$7.8 & 63$\pm$24 & 28$\pm$20 \\
 & Webster & 62$\pm$17 & 21$\pm$12 & 43$\pm$2.6 & 11$\pm$2.6 & \textbf{37$\pm$4.0}\smash{\textsuperscript{$\star$}} & \textbf{1.4$\pm$2.6}\smash{\textsuperscript{$\star$}} & 57$\pm$7.0 & 22$\pm$7.1 & 51$\pm$7.8 & 16$\pm$7.8 & 63$\pm$24 & 28$\pm$20 \\
 & SOTL & \textbf{59$\pm$9.5}\smash{\textsuperscript{$\dagger$}} & \textbf{21$\pm$9.5}\smash{\textsuperscript{$\ddagger$}} & 44$\pm$3.8 & 11$\pm$3.5 & 41$\pm$4.3 & 5.0$\pm$4.4 & 54$\pm$4.6 & 19$\pm$3.3 & 46$\pm$15 & 11$\pm$12 & 61$\pm$15 & 26$\pm$14 \\
 & MaxPressure & 77$\pm$21 & 35$\pm$17 & 42$\pm$5.9 & 9.3$\pm$5.1 & 45$\pm$10 & 9.4$\pm$8.6 & 59$\pm$10 & 25$\pm$10 & 45$\pm$8.3 & 9.7$\pm$8.0 & 67$\pm$20 & 33$\pm$17 \\
\hline
\multirow{4}{*}{RL} & IntelliLight & 124$\pm$25 & 69$\pm$24 & 41$\pm$5.1 & 8.8$\pm$3.9 & 40$\pm$3.2 & 3.6$\pm$2.5 & \textbf{52$\pm$6.9}\smash{\textsuperscript{$\ddagger$}} & \textbf{16$\pm$5.9}\smash{\textsuperscript{$\ddagger$}} & \textbf{43$\pm$5.3}\smash{\textsuperscript{$\ddagger$}} & \textbf{8.0$\pm$4.3}\smash{\textsuperscript{$\ddagger$}} & 65$\pm$18 & 29$\pm$16 \\
 & PressLight & 112$\pm$21 & 52$\pm$11 & 54$\pm$5.6 & 19$\pm$4.2 & 41$\pm$4.6 & 4.6$\pm$4.1 & 91$\pm$13 & 45$\pm$7.8 & 69$\pm$9.1 & 29$\pm$7.8 & 89$\pm$26 & 47$\pm$20 \\
 & AttendLight & 72$\pm$15 & 30$\pm$13 & \textbf{40$\pm$6.3}\smash{\textsuperscript{$\dagger$}} & \textbf{6.9$\pm$5.4}\smash{\textsuperscript{$\dagger$}} & 41$\pm$2.2 & 5.0$\pm$2.7 & 83$\pm$10 & 42$\pm$7.8 & 58$\pm$16 & 22$\pm$13 & 80$\pm$24 & 42$\pm$21 \\
 & UniTSA & 62$\pm$7.0 & 22$\pm$5.6 & \textbf{41$\pm$4.0}\smash{\textsuperscript{$\ddagger$}} & \textbf{7.8$\pm$3.7}\smash{\textsuperscript{$\ddagger$}} & 40$\pm$8.2 & 4.8$\pm$6.4 & 53$\pm$10 & 18$\pm$8.7 & \textbf{41$\pm$8.5}\smash{\textsuperscript{$\dagger$}} & \textbf{6.1$\pm$6.8}\smash{\textsuperscript{$\dagger$}} & \textbf{50$\pm$8.0}\smash{\textsuperscript{$\dagger$}} & \textbf{16$\pm$5.5}\smash{\textsuperscript{$\dagger$}} \\
\hline
\multirow{2}{*}{VLM} & VLMLight (32B) & 65$\pm$13 & 25$\pm$9.9 & 42$\pm$2.9 & 8.7$\pm$3.4 & 40$\pm$3.9 & 3.5$\pm$2.6 & \textbf{48$\pm$4.0}\smash{\textsuperscript{$\dagger$}} & \textbf{14$\pm$3.9}\smash{\textsuperscript{$\dagger$}} & 44$\pm$9.8 & 9.2$\pm$8.7 & \textbf{49$\pm$6.2}\smash{\textsuperscript{$\star$}} & \textbf{14$\pm$3.9}\smash{\textsuperscript{$\star$}} \\
 & VLMLight (3B) & 154$\pm$12 & 70$\pm$14 & 65$\pm$11 & 24$\pm$7.8 & 53$\pm$13 & 12$\pm$7.6 & 113$\pm$10 & 62$\pm$6.5 & 61$\pm$22 & 22$\pm$18 & 112$\pm$70 & 62$\pm$55 \\
\hline
Ours & \textbf{VLALight} (0.5B) & \textbf{57$\pm$13}\smash{\textsuperscript{$\star$}} & \textbf{17$\pm$9.8}\smash{\textsuperscript{$\star$}} & \textbf{38$\pm$1.7}\smash{\textsuperscript{$\star$}} & \textbf{6.6$\pm$2.2}\smash{\textsuperscript{$\star$}} & \textbf{38$\pm$1.4}\smash{\textsuperscript{$\ddagger$}} & \textbf{2.4$\pm$1.7}\smash{\textsuperscript{$\ddagger$}} & \textbf{44$\pm$4.2}\smash{\textsuperscript{$\star$}} & \textbf{9.6$\pm$3.3}\smash{\textsuperscript{$\star$}} & \textbf{36$\pm$4.2}\smash{\textsuperscript{$\star$}} & \textbf{3.0$\pm$2.7}\smash{\textsuperscript{$\star$}} & \textbf{55$\pm$16}\smash{\textsuperscript{$\ddagger$}} & \textbf{20$\pm$13}\smash{\textsuperscript{$\ddagger$}} \\
\hline
\end{tabular}
\caption{Per-city comparison on the regular flow (mean$\pm$std over 4 seeds, seconds). The top three per column are \emph{boldfaced} and ranked $^\star$/$^\dagger$/$^\ddagger$.}
\label{tab:comparison_summary}
\end{table*}

\subsection{Experimental Settings}

\subsubsection{Datasets and Simulation}

We evaluate VLALight on the six real-world intersections introduced in the Method section, rendered in TransimHub~\cite{wang2025transimhub}; Figure~\ref{fig:intersections} pairs each real intersection with its simulated counterpart. A single model is trained and evaluated across all sites in closed-loop simulation, on the \emph{regular flow} used to collect the expert demonstrations and three \emph{test flows} (\emph{off-peak}, \emph{rush-hour}, and \emph{fluctuating}), with four independent seeds per city-flow pair; full dataset details are given in the supplementary material (Section~B.1).

\begin{center}
\includegraphics[width=\columnwidth]{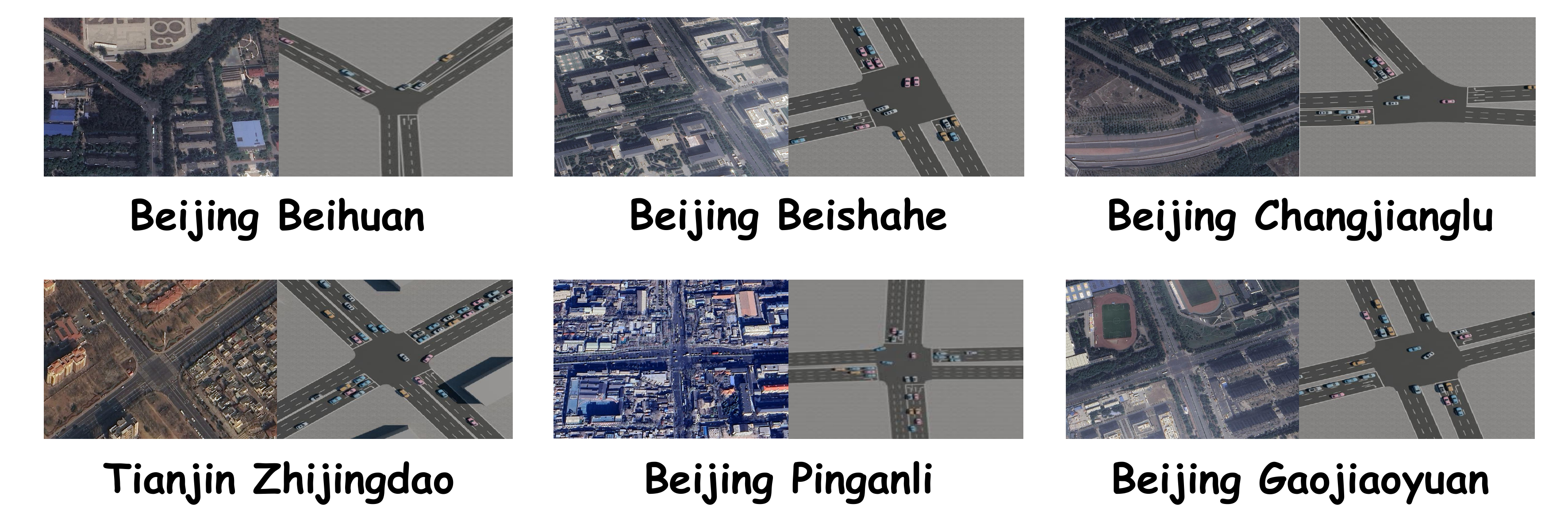}
\captionof{figure}{The six real-world intersections in our dataset. For each site, the satellite view of the real intersection (left) is paired with its rendered simulation (right).}
\label{fig:intersections}
\end{center}

\subsubsection{Evaluation Metrics}

We report the mean and standard deviation of four metrics: Average Travel Time (ATT) and Average Waiting Time (AWT) for overall traffic efficiency, and Emergency Travel Time (ETT) and Emergency Waiting Time (EWT) for emergency-vehicle service, all in seconds (lower is better).

\subsubsection{Compared Methods}

We compare against nine controllers from three families: four rule-based methods, namely FixTime, Webster~\cite{webster1958traffic}, SOTL~\cite{gershenson2005self}, and MaxPressure~\cite{varaiya2013maxpressure}; four RL-based methods, namely IntelliLight~\cite{wei2018intellilight}, PressLight~\cite{wei2019presslight}, AttendLight~\cite{oroojlooy2020attendlight}, and UniTSA~\cite{wang2024unitsa}; and the VLM-based VLMLight~\cite{wang2025vlmlight}, whose fast branch is a pretrained UniTSA/PPO controller and whose slow branch invokes cloud-scale VLM/LLM agents. VLMLight is evaluated under the same protocol with two backbone sizes: the cloud configuration 32B and a 3B variant. VLALight uses a 0.5B backbone and runs entirely locally. Baseline configurations are detailed in the supplementary material (Section~B.3).

\subsection{Traffic Efficiency Comparison}

\begin{figure}[t]
\centering
\includegraphics[width=\columnwidth]{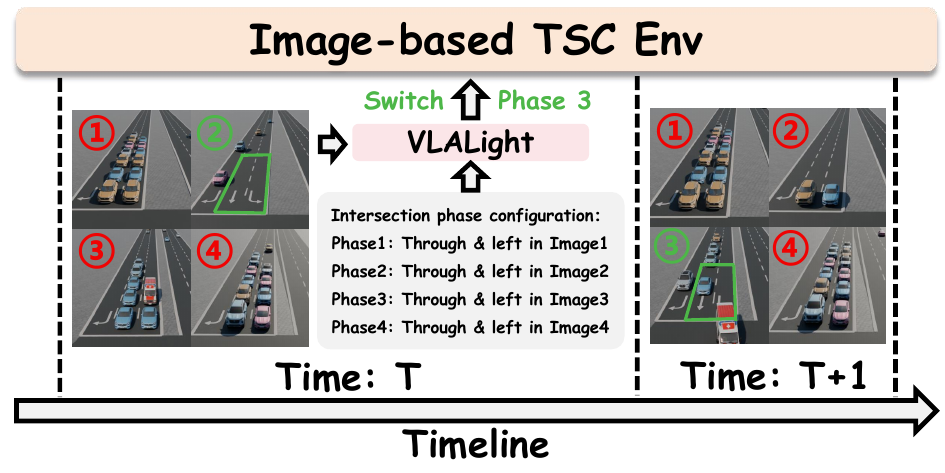}
\caption{An emergency-priority example in TransimHub. At time $T$ an emergency vehicle approaches the intersection in one camera view; VLALight switches to the phase 3, and at time $T{+}1$ the vehicle has crossed the intersection.}
\label{fig:evcase}
\end{figure}

Table~\ref{tab:comparison_summary} reports the per-city comparison under the regular flow; complete breakdowns of all four metrics, organized by intersection and by traffic flow, are provided in the supplementary material (Section~C), and generalization to the test flows is analyzed below. VLALight provides the best emergency-vehicle service of all compared methods: it attains the lowest ETT and EWT in four of the six cities and, pooled over all six cities, improves on the cloud-scale VLMLight (32B) by $7.0\%$ in ETT and $21.1\%$ in EWT. The two exceptions are Beijing Changjianglu and Tianjin Zhijingdao, where the rule-based controllers or VLMLight (32B) rank ahead. Figure~\ref{fig:evcase} shows a concrete example: when an emergency vehicle approaches in one camera view, VLALight switches to the phase serving its movement at the next decision step, letting the vehicle pass without stopping; an extended case study is provided in the supplementary material (Section~D). These results show that an end-to-end 0.5B policy can match or surpass the emergency-vehicle service of a cloud-scale cascaded system.

\subsection{Inference Latency and Deployment Analysis}

Beyond control quality, real-time deployment hinges on inference latency, which Table~\ref{tab:latency} summarizes. In routine steps VLMLight can fall back on its fast RL branch, but emergency-aware service is exactly where its slow branch fires: the cascaded design then chains several agent calls for a single decision, costing $10.9$~s with the cloud 32B backbone and $4.2$~s on average with a locally deployed 3B backbone, both close to or beyond the $5$~s decision interval. VLALight instead outputs the action in a single local forward pass, averaging $224$~ms per decision on one RTX 4090 GPU, which supports fully on-premises deployment without streaming surveillance feeds to the cloud. VLALight is therefore the only compared controller that stays comfortably within the decision budget with emergency awareness always enabled.

\begin{center}
\small
\setlength{\tabcolsep}{2.5pt}
\begin{tabular}{@{}l|c|c@{}}
\hline
Method & Model size & Latency/decision \\
\hline
VLMLight & 32B (cloud) & 10.9 s \\
VLMLight & 3B (local) & 4.2 s \\
\textbf{VLALight} & \textbf{0.5B} (local) & \textbf{0.2 s} \\
\hline
\end{tabular}
\captionof{table}{Inference latency per control decision in emergency scenarios, where VLMLight's slow branch fires.}
\label{tab:latency}
\end{center}

\subsection{Ablation Study}

\begin{figure}[t]
\centering
\includegraphics[width=\columnwidth]{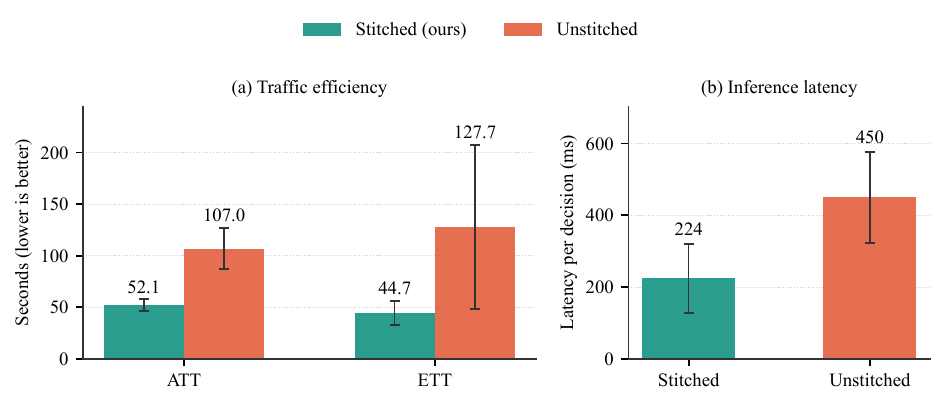}
\caption{Ablation of the visual-input design: stitched vs.\ unstitched visual input. (a) Pooled ATT and ETT under the regular flow; (b) inference latency per decision.}
\label{fig:ablation}
\end{figure}

As Figure~\ref{fig:ablation} shows, we ablate the visual-input design by comparing our single stitched image against feeding the four camera views as four separately encoded images, which quadruples the visual-token count from $256$ to $1{,}024$. The single stitched input wins in every city and every seed under the regular flow, reducing pooled ATT from $107.0$~s to $52.1$~s and ETT from $127.7$~s to $44.7$~s. The quadrupled tokens inflate activation memory and limit the training batch without adding spatial information, and a compact 0.5B backbone struggles to relate separately encoded views through self-attention. The variant is also twice as slow at inference, at $450$~ms per decision compared with $224$~ms for the stitched input. The stitched input is thus superior in control quality, emergency response, memory footprint, and latency alike, which validates encoding the directional views jointly as one image. An ablation of the minimum-green dwell constraint is provided in the supplementary material (Section~C).

\subsection{Generalization Analysis}

We assess generalization along two axes: unseen traffic-flow patterns and unseen intersection topologies.

\paragraph{Generalization to unseen traffic flows.}

\begin{figure}[b]
\centering
\includegraphics[width=\columnwidth]{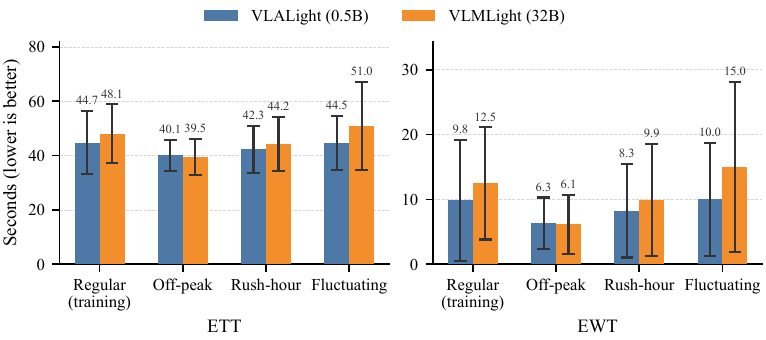}
\caption{Generalization to unseen traffic flows: ETT and EWT on the regular flow and the three test flows.}
\label{fig:flow_generalization}
\end{figure}

\begin{figure}[t]
\centering
\includegraphics[width=0.9\columnwidth]{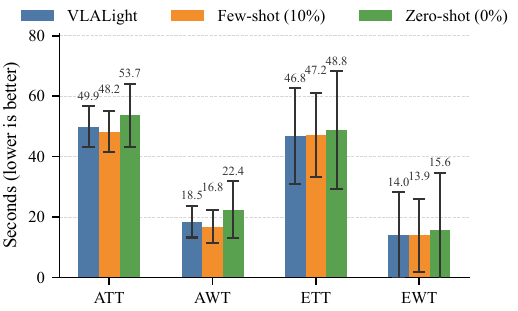}
\caption{Generalization to unseen intersection topologies: transfer to two unseen intersections with few-shot and zero-shot adaptation.}
\label{fig:generalization}
\end{figure}

We further evaluate the same model, without any retraining, on the three test flows (\emph{off-peak}, \emph{rush-hour}, and \emph{fluctuating}; per-flow results in the supplementary material, Section~C.3). Pooled over the three test flows, VLALight retains the best emergency service, improving on VLMLight (32B) by $5.8\%$ in ETT and $20.8\%$ in EWT, a margin close to that on the regular flow. As Figure~\ref{fig:flow_generalization} shows, the advantage grows as conditions depart from the training distribution: on par under the \emph{off-peak flow}, $16.5\%$ lower EWT under the \emph{rush-hour flow}, and $33.1\%$ under the \emph{fluctuating flow}. The policy thus transfers well to demand patterns it never encountered during training.

\paragraph{Generalization to unseen topologies.}
We split the six intersections into four training sites and two unseen test sites, Beijing Beishahe and Tianjin Zhijingdao, and Figure~\ref{fig:generalization} compares three variants on the two test sites: the fully trained model, a few-shot variant fine-tuned with only 10\% of the target intersections' data, and a zero-shot variant given no target data at all. The few-shot variant stays within $1$--$3\%$ of the fully trained model on ETT and EWT, and even slightly improves ATT and AWT. Without any target data, the zero-shot variant remains competitive: its ETT and EWT rise by only $4.3\%$ and $11.4\%$, and ATT by $7.6\%$. VLALight thus learns transferable visual-language priors rather than memorizing the training intersections.

\section{Conclusion}

We presented VLALight, a lightweight end-to-end vision-language-action framework for intersection traffic signal control. By stitching multiple directional camera views into a unified visual input grounded by textual instructions, a compact 0.5B model maps intersection observations directly to discrete signal actions in a single forward pass. Experiments on six real-world intersections under four traffic-flow patterns show that VLALight attains the strongest emergency-vehicle service among all compared controllers, with the lowest ETT and EWT in four of the six cities and a $21.1\%$ lower pooled EWT than the cloud-scale VLMLight, while answering each decision in $224$~ms on local hardware and generalizing to unseen intersection topologies and traffic-flow patterns.

Several directions remain open for future work, which also mark the current limitations of VLALight: the policy is trained by behavior cloning of simulation-collected expert labels, and reinforcement fine-tuning is a natural next step; decisions are made from single frames, where temporal memory could help; and evaluation is limited to simulation, leaving real-world validation and formal safety guarantees to future studies.

{\small\bibliography{aaai2027}}

\end{document}